\documentclass{scspaperproc}
\usepackage{latexsym}
\usepackage{graphicx}
\usepackage{mathptmx}
\usepackage{amsmath}
\usepackage{amsfonts}
\usepackage{amssymb}
\usepackage{amsbsy}
\usepackage{amsthm}
\usepackage{array}
\usepackage[dvipsnames]{xcolor}
\usepackage[table]{xcolor}
\usepackage{makecell}
\usepackage{wrapfig}
\usepackage{comment}
\usepackage{tikz}
\usepackage{booktabs}
\usepackage{pifont}
\usepackage{multirow}
\usepackage{siunitx}
\usepackage{longtable}
\usepackage{eucal}

\usetikzlibrary{positioning}
\usetikzlibrary{arrows.meta}
\usetikzlibrary{shapes.geometric}
\newcolumntype{x}[1]{>{\centering\let\newline\\\arraybackslash\hspace{0pt}}p{#1}}

\newcommand{\s}{~\phantom{\texttimes}}%
\newcommand{\hdr}[1]{\multicolumn{1}{c}{\textbf{#1}}}
\newcommand{\spl}{\ensuremath{\!~{\scriptstyle\pm}~\!}} % tighter, smaller ±
\newcommand{\y}{\raisebox{0.3ex}{\footnotesize$\checkmark$}}
\newcommand{\n}{\raisebox{0.3ex}{\footnotesize$\times$}}
\newcommand{\block}[1]{}

\usepackage[pdftex,colorlinks=true,urlcolor=black,citecolor=black,anchorcolor=black,linkcolor=black,bookmarks=false]{hyperref}
\begin{document}
\SCSpagesetup{McClurg and Wagner}
\def\SCSconferencename{Annual Modeling and Simulation Conference}
\def\SCSconferenceacro{ANNSIM'26}
\def\SCSpublicationyear{2026}
\def\SCSconferenceeditors{G. Rabadi, V. Prabhu, R. Cárdenas, A. Bany Abdelnabi, J. Jabbour, and M. Germanos}
\def\SCSconferencedates{May 4-7}
\def\SCSconferencevenue{University of Central Florida, Orlando, Florida, USA}
\title{Developing a Discrete-Event Simulator\\of School Shooter Behavior from VR Data}
\author[\authorrefmark{1}]{Christopher A. McClurg}
\author[\authorrefmark{1}]{Alan R. Wagner}
\affil[\authorrefmark{1}]{Department of Aerospace Engineering, Pennsylvania State University}

\maketitle

% ============================================ 

\section*{Abstract}
Virtual reality (VR) has emerged as a powerful tool for evaluating school security measures in high-risk scenarios such as school shootings, offering experimental control and high behavioral fidelity. However, assessing new interventions in VR requires recruiting new participant cohorts for each condition, making large-scale or iterative evaluation difficult. These limitations are especially restrictive when attempting to \textit{learn} effective intervention strategies, which typically require many training episodes. To address this challenge, we develop a data-driven discrete-event simulator (DES) that models shooter movement and in-region actions as stochastic processes learned from participant behavior in VR studies. We use the simulator to examine the impact of a robot-based shooter intervention strategy. Once shown to reproduce key empirical patterns, the DES enables scalable evaluation and learning of intervention strategies that are infeasible to train directly with human subjects. Overall, this work demonstrates a high-to-mid fidelity simulation workflow that provides a scalable surrogate for developing and evaluating autonomous school-security interventions.

\textbf{Keywords:} Active Shooter, Virtual Reality, Discrete-Event Simulation, Reinforcement Learning 

% ============================================ 
\section{Introduction}\label{sec:int}
% ============================================ 

School shootings remain a persistent and growing concern in the United States, as more than half (53.3\%) of documented gun-related school incidents occurred within the past seven years~\cite{riedman2025k}. Exposure to such events is associated with elevated rates of post-traumatic stress disorder (PTSD) and anxiety among students~\cite{suomalainen2011controlled, elklit2013psychological}. Although numerous school-security measures have been implemented---ranging from hardened infrastructure to behavioral profiling---many lack empirical validation of their efficacy~\cite{addington2009, schwartz2016role, johnhopkins2016}. In addition, some measures have been linked to increased student anxiety, reduced trust, and erosion of school climate~\cite{bachman2011predicting}.

Evaluating school-security measures for school shootings is difficult, as the detailed behavioral data required for validation are limited and cannot be systematically collected from real events~\cite{briggs2016active}. Although ecologically valid, unannounced experiments in real schools would pose unacceptable physical, ethical, and psychological risks~\cite{suomalainen2011controlled, elklit2013psychological}. Virtual reality (VR) may offer a practical alternative, allowing participants to assume the role of a shooter in a simulated school environment while detailed behavior and physiological measurements are taken~\cite{mcclurg2025using}. Prior work has demonstrated statistical equivalence between VR-generated behavior and real shooter data while also showing promise for evaluating intervention strategies~\cite{mcclurg2025robot}.

Despite its advantages, conducting human-subject experiments in VR to evaluate school-security measures does not scale. Each change in experimental condition---whether testing a new intervention or refining an existing one---requires a separate human-subject study with a new participant cohort. To address this limitation, this paper proposes a discrete-event simulation (DES) as a scalable surrogate for conditional human-subject experiments, in which shooter behavior is modeled as a stochastic process learned from human-subject data in VR. 

The remainder of the paper is organized as follows. Section~\ref{sec:background} reviews the relevant background. Section~\ref{sec:data_description} describes the participant data collected in virtual reality that was used to construct and calibrate the simulator. Sections~\ref{sec:model_description}--\ref{sec:model_validation} then describe and evaluate the fidelity of the discrete-event simulator. Section~\ref{sec:policy_demo} demonstrates the use of the simulator in a sample-intensive agent-learning task, which would otherwise be infeasible to carry out using a series of human-subject experiments.

% ============================================
\section{Background}\label{sec:background}
% ============================================ 
A school-security intervention refers to any measure intended to improve safety within a school environment, including those aimed at reducing the likelihood or severity of shooting incidents. For many proposed interventions, an autonomous system (e.g., an alerting system, a surveillance platform, or a mobile robot) must determine how to act in response to a shooter’s behavior. Developing or learning such a strategy requires a way to represent and model agent decision-making. Accordingly, this section reviews the notion of an agent policy, the mechanisms for learning policies, and the role of surrogate simulation in supporting sample-intensive learning. We use the term \textit{agent} here to generalize across the different types of autonomous systems that could be used to improve school safety in these scenarios.

\subsection{Agent Policies}
A \textit{policy} maps an agent’s state or observation to an action~\cite{bellman1957dynamic, puterman1994mdp}. In agent-based modeling (ABM), a common framework for studying active-shooter incidents, shooter policies are typically implemented as hand-crafted behavioral rules. Examples include remaining stationary~\cite{briggs2016active}, wandering randomly~\cite{anklam2014mitigating}, or moving toward the nearest civilian~\cite{hayes2014agent, stewart2017active, lee2018agent, lee2019agent}. These rules are executed at fixed time-step intervals (typically 1--10~Hz), causing temporal patterns to emerge because of repeated rule execution.

Discrete-event simulation (DES) provides an alternative way to represent agent policies~\cite{tocher1963art}. Rather than selecting actions at fixed time intervals, time advances directly to the next meaningful event, such as a shooter entering a new room or the arrival of first responders. This event-driven structure naturally captures variable-duration behaviors and supports policies defined through data-driven transitions and stochastic event outcomes. As a result, DES provides a flexible and empirically-grounded alternative to traditional rule-based ABM approaches for modeling shooter behavior.

\subsection{Learning Policies}
Ideally, an intervening agent’s policy could be learned from experience rather than specified in advance. An agent tasked with preventing a shooting, for example, may have the ability to intervene but must learn how and when to act in response to dynamic shooter behavior. Reinforcement learning provides a common approach for this, allowing an agent to improve its behavior through repeated interaction with an environment by selecting actions, observing outcomes, and receiving feedback in the form of a reward~\cite{sutton1998introduction}. Reinforcement learning is, however, typically sample-intensive, often requiring a large number of trials before effective policies are learned. The approach becomes impractical when policy learning requires human-subject experiments (e.g., VR-based shooter role-play) to generate each variation of the experimental conditions. Hence, the primary goal of this paper is to develop a method that uses previously collected data to model the shooter's behavior in a manner that enables scalable learning of agent intervention policies without the need for iterative human-subject data collection.

\subsection{Surrogate Modeling}
Because direct policy learning in high-fidelity environments is often impractical, surrogate environments are commonly used to support scalable policy learning. The most relevant body of work in this space is simulation-to-reality (sim-to-real) research~\cite{salvato2021crossing, zhao2020sim}, which has largely focused on robotics. In sim-to-real settings, policies are learned in simulation and later deployed on physical robots, creating a well-known \textit{reality gap} when the simulator does not adequately represent the real system. To mitigate this gap, researchers have developed techniques such as domain randomization, which introduces controlled variability into simulator parameters to promote robustness across environmental conditions~\cite{tobin2017domain, james2019sim, peng2018sim}, and domain adaptation, which aligns simulated and real observation or action spaces using learned mappings or shared representations~\cite{hanna2017grounded, ghadirzadeh2017deep, rusu2017sim}. Hybrid approaches that combine both strategies further improve transferability and reduce reliance on costly real-world data collection~\cite{van2019sim, tan2018sim, hu2021sim}.

Although these examples originate in physics-based robotics simulators, the underlying multi-fidelity principles generalize. In the present context, participant data from virtual reality provides empirically-grounded shooter behavior data but is too costly for the repeated trials required for policy learning. A discrete-event simulator derived from participant data serves as a mid-fidelity surrogate: it approximates school shooter behavior while enabling large-scale, low-cost experimentation. Moreover, the stochastic variability inherent in the participant data naturally induces a form of domain randomization, exposing learned policies to a distribution of plausible human behaviors.

% ============================================
\section{Data Description}\label{sec:data_description}
% ============================================

Our surrogate simulator is calibrated using behavior data from two VR studies~\cite{mcclurg2025using, mcclurg2025robot}, in which participants role-played as an active shooter in a high-fidelity reconstruction of Columbine High School. Participants navigated the environment using a foot interface (Cybershoes) while aiming and firing with VR hand controllers. The environment was populated with non-player characters (NPCs) following a “Run. Hide. Fight.” protocol, along with two mobile robots attempting to intervene and slow down the shooter. The complete dataset consists of 210 five-minute episodes logged at 2~Hz. Each episode includes the position of participants and robots, as well as NPC state information (e.g., alive or dead) and records of shots fired. For the remainder of this paper, we refer to this data as the \textit{participant data}.

In the present work, we focus on a subset of the participant data, namely the no-robot and robot-with-smoke conditions, each including 60 episodes. The no-robot condition did not include a robot intervention; consequently, only shooter information was collected. The robot-with-smoke condition included a robot intervention in which the robot deployed smoke to confuse and delay the shooter. To prepare the data for discrete-event simulation, the environment was spatially discretized into \textit{regions} (see Fig.~\ref{fig:map_to_graph}, left), and transitions between regions were treated as discrete \textit{events}. Event outcomes were extracted from the 2~Hz simulation logs across all participants. Whenever a participant entered a new region, cumulative outcomes associated with the previous region---time spent, shots fired, and victims---were recorded. Aggregating these outcomes across participants produced region-level distributions. Regions were further grouped into semantically homogeneous categories (classrooms, hallways, large common areas, stairwells, entrances, and outdoor regions), allowing outcomes to be pooled at the group level. Aggregating across all regions yielded global distributions.

% ============================================
\section{Model Description} \label{sec:model_description}
% ============================================

Our discrete-event simulator consists of three primary components: (1) a shooter-transition model that governs movement between regions of the school layout, (2) a shooter-event model that characterizes dwell time, shots fired, and victims within each region visited, and (3) a robot-effects model that modulates event outcomes based on robot presence and smoke intensity. Together, these components generate a stochastic, temporally grounded event sequence derived from participant behavior. The subsections that follow describe each component. To support reproducibility, we provide a public implementation of the modeling framework online on GitHub,\footnote{\url{https://github.com/chrismcclurg/vr-shooter-des}} including code for region-graph construction, transition-model training, event-sampling routines, and robot-effects modeling.

\subsection{Shooter Transitions} \label{sec:model_trans_comp}

\begin{figure}[!h]
  \centering
  \includegraphics[width=1.0\linewidth]{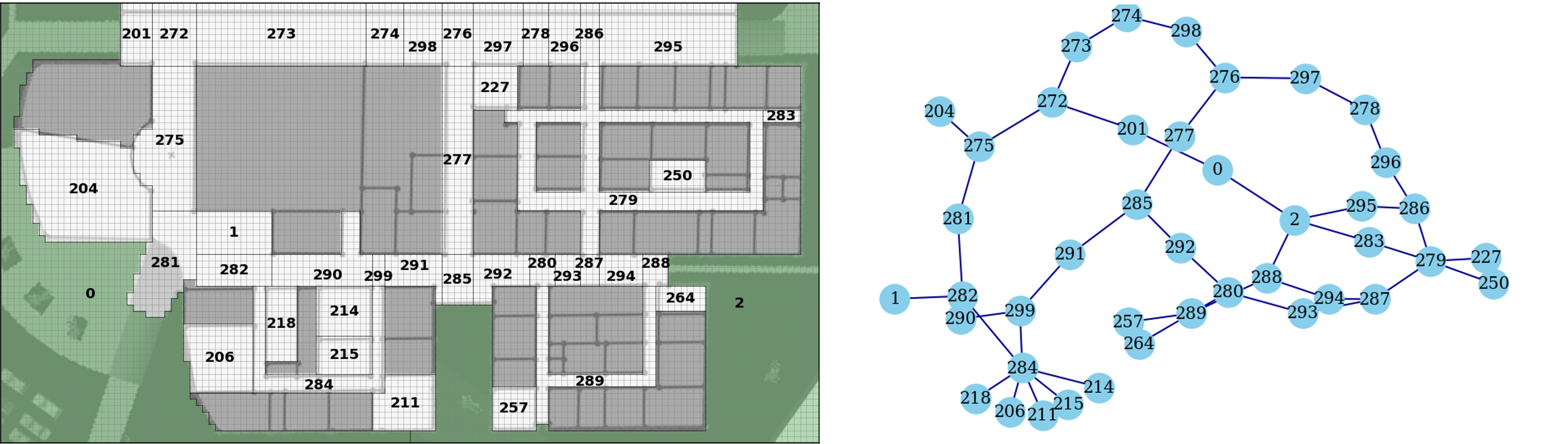}
  \caption{The annotated floor plan (left) was transformed into a graph representation (right), with regions represented as nodes and edges defining their adjacencies. Shooter behavior was modeled as a sequence of events, each defined by the time spent within a region.}
  \label{fig:map_to_graph}
\end{figure}

Shooter movement between building regions was modeled using a graph neural network (GNN)~\cite{scarselli2008graph}. The school layout was represented as a directed graph, with regions connected by edges indicating feasible movements (see Fig.~\ref{fig:map_to_graph}, right). Each region was associated with a feature vector capturing local spatial and contextual information. Given the current region, the GNN predicted a probability distribution over neighboring regions. Models were trained using the participant data (Sec.~\ref{sec:data_description}), which was first randomized by episode and split into training, validation, and test sets. Once trained, the GNN transition model was used within the discrete-event simulator to generate rollouts of shooter movement.

We used a three-layer GraphSAGE~\cite{hamilton2017inductive} model to learn representations of each region based on its connections and attributes. The resulting representations were fed into a two-layer classifier that estimated the likelihood of moving to each neighboring region. These likelihoods were normalized to form a probability distribution, which the simulator used to either sample a transition or select the most likely next region. Models were implemented in TensorFlow/Keras. GraphSAGE layers used 64 hidden units with ELU activations, $\ell_2$ regularization ($10^{-4}$), and dropout (0.1). Training used Adam with a learning rate of $10^{-3}$, batch size 32, early stopping (patience 15), and learning-rate reduction on plateau. 

\begin{figure}[h!]
\centering
\includegraphics[width=0.9\linewidth]{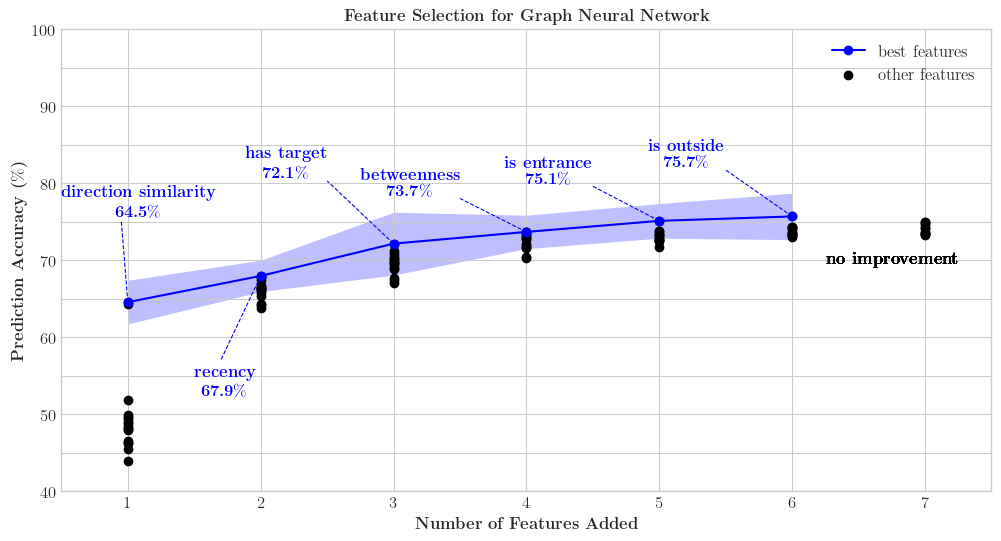}
\caption{Greedy forward-selection results. Blue markers show the feature added at each stage of the selection process, with mean prediction accuracy and standard deviation across five folds. Black points represent the accuracies obtained by all other intermediate combinations evaluated during the search.}
\label{fig:greedy_search}
\end{figure}

At each transition decision, a feature vector was computed for each possible next region. Features were selected from three categories: semantic room attributes, graph-theoretic measures, and dynamic state-dependent descriptors. A greedy forward-selection procedure identified the six features that produced the largest improvement in cross-validated prediction accuracy (Figure~\ref{fig:greedy_search}). The selected features were:
\begin{itemize}
    \item \texttt{direction\_similarity}: cosine similarity between the direction of the previous movement and the direction toward each neighboring region.
    \item \texttt{recency}: normalized time since the shooter last visited the region (larger values indicate more recent visits).
    \item \texttt{has\_target}: binary indicator of whether the region currently contains potential targets.
    \item \texttt{betweenness}: graph centrality measure reflecting how often the region lies on shortest paths between other regions.
    \item \texttt{is\_entrance}: binary indicator of whether the region is a building entrance.
    \item \texttt{is\_outside}: binary indicator of whether the region corresponds to an outdoor location.
\end{itemize}

\subsection{Shooter Events}
\begin{figure}[!t]
\centering
\includegraphics[width=1.0\linewidth]{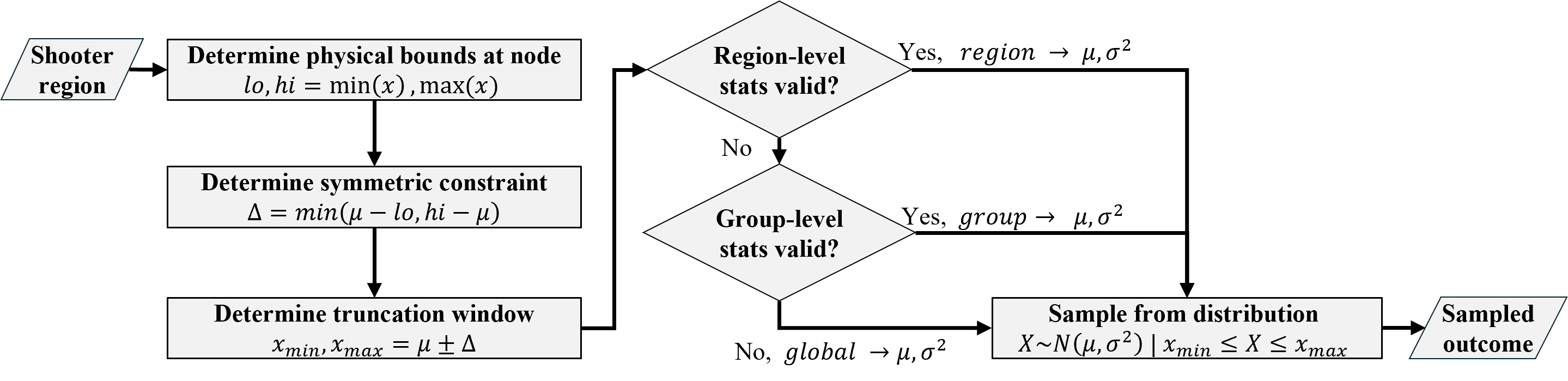}
\caption{Flowchart of the hierarchical truncated-normal sampling procedure. Event outcomes are sampled from a truncated normal distribution parameterized by region-, group-, or global-level statistical moments, depending on data availability.}
\label{fig:sampling_flow}
\end{figure}

Shooter events model what occurs while the shooter remains within a region---including time spent, shots fired, and victims---by treating each visit as a single event with jointly generated outcomes. As described in Sec.~\ref{sec:data_description}, these outcomes were derived from participant data and aggregated at the region, group, and global levels. During simulation, event outcomes are sampled conditional on the current region using a hierarchical truncated-normal method (Figure~\ref{fig:sampling_flow}) that matches observed first and second moments of the data, while enforcing physical limits. Time was required to be positive and capped by the remaining episode duration; shots were positive with no effective upper bound; and victims were positive and capped by the maximum observed values for the region.

The sampling procedure proceeds in two stages. First, a \textit{moment-matched truncation} step constructs a truncated normal distribution whose post-truncation mean and variance match the region-level moments. The truncation interval is defined symmetrically about the empirical mean, with its half-width chosen as the largest value within the physically feasible bounds for the metric. Symmetric truncation ensures that the truncated mean and variance vary consistently with the underlying (pre-truncation) standard deviation, yielding stable and identifiable moment matching. In contrast, asymmetric truncation can lead to unstable estimates under sparse or highly skewed data. Second, a hierarchical fallback step applies the same moment-matching procedure at coarser levels when region-level data are limited. Specifically, group-level and then global-level moments are used.

The sampling procedure is supported by the structure of the empirical distributions. Region-level outcomes were typically unimodal: Gaussian kernel density estimation (KDE) applied to each region distribution revealed a median of 1.0 peak for time, shots, and victims. The distributions also exhibited varying degrees of skewness---time and shots were right-tailed (median skewness of 1.30 and 3.01, respectively), whereas victim counts were nearly symmetric (median skewness of 0.17). Although such skewness would ordinarily argue against normal sampling, truncating the distribution within physical bounds preserves empirical means and variances while suppressing unstable tail behavior arising from sparse or highly skewed observations.

\subsection{Robot Effects}\label{sec:robot_effects}

Robot effects model how a robot’s presence perturbs event outcomes through a spatial influence that depends on its proximity and activity. During simulation, robot influence acts as a scalar modifier on event outcomes based on smoke intensity within the current region. For each event outcome $X_i$ (time, shots, or victims) at region $i$, the baseline value is adjusted according to local robot influence, yielding

\begin{equation}
    \bar{X}_i(t) = X_i + R_i(t) k_{x,i},
\end{equation}

\noindent where $X_i$ denotes the baseline event outcome, $R_i(t)$ quantifies robot influence, and $k_{x,i}$ is an outcome-specific coefficient of robot effect. The coefficient $k_{x,i}$ was estimated from robot-present episodes using shrinkage-weighted linear regression on region-specific residuals, allowing effects to vary across regions while stabilizing estimates when data are limited. Robot influence is computed using a graph-based smoke model. Let $S_j(t)$ denote the normalized smoke intensity at region $j$, estimated from robot occupancy over a trailing time window, and let $J$ denote the set of all regions. Smoke influence decays with shortest-path distance according to

\begin{equation}
    w_{ij} = e^{-\lambda D_{ij}},
\end{equation}

\noindent where $D_{ij}$ is the shortest-path distance between regions $i$ and $j$ and $\lambda$ is a decay parameter. The resulting influence at region $i$ is computed as

\begin{equation}
    R_i(t) = \alpha(t)\sum_{j \in J} S_j(t)\, w_{ij},
\end{equation}

\noindent where $\alpha(t)\in[0,1]$ is a saturating emission term that increases with cumulative robot presence time, representing temporal smoke accumulation. This formulation captures both spatial diffusion and temporal accumulation of smoke, allowing empirical robot effects---whether increasing dwell time or reducing shot and victim rates---to be reproduced consistently within the simulator.

% ============================================
\section{Model Evaluation}\label{sec:model_validation}
% ============================================

This section evaluates each component of the model. For the shooter-transitions component, we evaluate one-step prediction accuracy against several baselines. For the shooter-events and robot-effects components, we compare synthetic outcome distributions with empirical data. Because these components are stochastic, results are computed over repeated independent rollouts and summarized using sample means and standard deviations. Statistical tests and, where appropriate, confidence intervals are used to assess variability and agreement with the empirical data.

\subsection{Shooter Transitions}
To evaluate shooter-transition prediction, we compared the learned model to several heuristic baseline methods from the literature. Drawing from prior agent-based modeling research~\cite{hayes2014agent, stewart2017active, lee2018agent, lee2019agent}, we implemented random (RA) and closest-target (CT) movement rules, which select the next region either uniformly at random or by minimizing distance to the nearest NPC target. A constant-velocity (CV) rule, commonly used in human-trajectory prediction~\cite{pellegrini2009you, alahi2016social}, was adapted to the discrete-region setting by selecting the neighboring region whose centroid direction most closely aligns with the previous transition. We also included graph-based heuristics that direct movement toward or away from the nearest entrance region (CE / FE). Finally, a region-based heuristic assumed shooters preferentially move toward physically larger areas (LA). These baselines were compared to our proposed graph neural network (GNN) model, described in Section~\ref{sec:model_trans_comp}.

\begin{figure}[!b]
\centering
\includegraphics[width=1.0\linewidth]{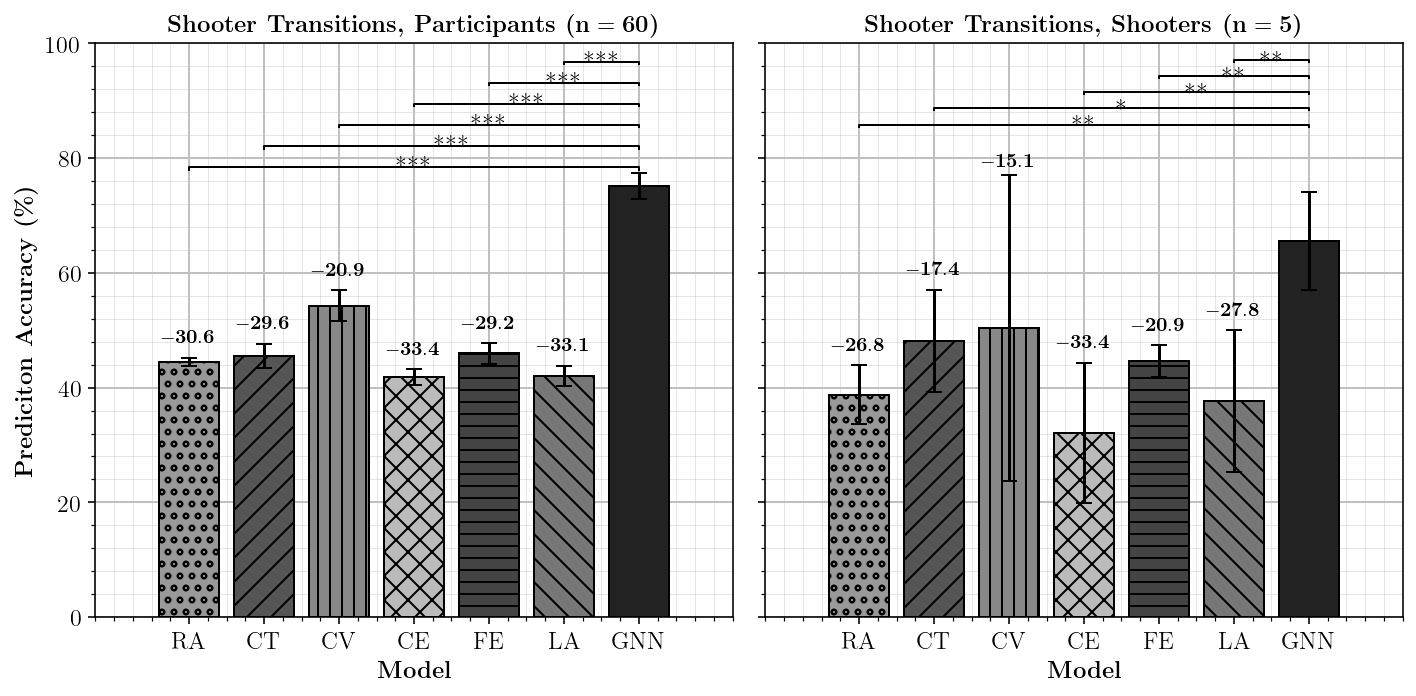}
\caption{Prediction accuracy of region-transition models evaluated on participant data (left) and shooter data (right). Bars indicate mean accuracy with 95\% confidence intervals. Bold labels above each bar report the absolute difference relative to the GNN model in percentage points. Asterisks denote significance from Welch’s unequal-variance $t$-test: *$p < 0.05$, **$p < 0.01$, ***$p < 0.001$.}
\label{fig:bar-plot_transitions}
\end{figure}

We evaluated transition-prediction performance using the following cross-validation protocol. The participant data were randomly ordered by episode and partitioned into five subsets. A leave-one-subset-out cross-validation scheme was then applied, with four subsets (80\%) used for training and the remaining subset (20\%) held out for testing. This procedure was repeated until each subset had served as the test set once. In addition, transitions from five real shooters were included as an external test set. These trajectories were extracted from publicly available case reports and manually annotated into discrete region sequences aligned with the corresponding school layouts. Prediction errors were aggregated at the episode level, yielding one independent error value per participant or shooter. The resulting effective sample sizes for statistical testing were $n=60$ for the VR participant set and $n=5$ for real-shooter cases.

The results of the shooter-transition evaluation are shown in Figure~\ref{fig:bar-plot_transitions}. For previously unseen participants ($n=60$), the GNN achieved significantly higher next-region prediction accuracy than all baseline methods. Welch’s unequal-variance $t$-tests confirmed that the GNN outperformed every baseline ($p<0.001$). Performance on real-shooter data ($n=5$) exhibited the same qualitative pattern. The GNN again achieved the highest accuracy and significantly outperformed each baseline according to Welch’s $t$-tests ($p<0.05$). Together, these results indicate that the learned transition model generalizes not only to held-out participant data but also to externally sourced out-of-distribution shooter trajectories.

% ---------------------------------------------
\subsection{Shooter Events}

To evaluate shooter-event prediction, we compared generated outcomes to the observed sequences of events of each participant. The evaluation focused on two criteria: matching marginal outcome statistics and preserving realistic spatial and temporal structure within simulated rollouts. To assess distribution accuracy, differences in means and variances between generated and participant outcomes were tested using Welch’s unequal-variance $t$-tests and Levene’s tests, respectively. Spatial fidelity was quantified using Jensen--Shannon divergence (JSD), with smaller values indicating greater distributional similarity. Temporal fidelity was assessed using Spearman’s rank correlation (SRC) by comparing the participant correlation between dwell time and event outcomes (shots and victims) to the corresponding correlation in generated data.

\begin{table}[!b]
\centering
\small
\setlength{\tabcolsep}{7pt}

\caption{Generated event outcomes compared to observed no-robot participant data. The numbers of generated and observed samples were 600 and 60, respectively. See notes for moment matching indicators.}
\label{tab:episode_outcomes}
\begin{tabular}{ll
    S[table-format=2.1]
    S[table-format=3.1]
    S[table-format=3.1]
    S[table-format=2.1]}
\toprule
\multicolumn{2}{c}{} &
\multicolumn{4}{c}{\textbf{Episode Outcomes: M ± SD with Moment-Matching Indicators}} \\
\cmidrule(lr){3-6}
\textbf{Pooling} & \textbf{Variant}
  & \hdr{Nodes}
  & \hdr{Time$^\dagger$}
  & \hdr{Shots}
  & \hdr{Victims} \\
\midrule

%% EMPIRICAL
\multicolumn{1}{c}{---}
  & \textit{Participants}
  & {\cellcolor{Gray!20}38.6\spl14.9~\s\s}
  & {\cellcolor{Gray!20}299.5\spl0.2~\s\s}
  & {\cellcolor{Gray!20}93.7\spl42.3~\s\s}
  & {\cellcolor{Gray!20}33.0\spl15.5~\s\s} \\
\addlinespace

%% GLOBAL MODELS
Global
  & Means
  & {\cellcolor{ForestGreen!20}38.6\spl14.9~\y\y}
  & {\cellcolor{Goldenrod!20}303.9\spl122.2~\y\n}
  & {\cellcolor{ForestGreen!20}86.6\spl41.9~\y\y}
  & {\cellcolor{Goldenrod!20}30.1\spl10.3~\y\n} \\

  & Sampling
  & {\cellcolor{ForestGreen!20}38.6\spl14.9~\y\y}
  & {\cellcolor{Goldenrod!20}301.6\spl123.1~\y\n}
  & {\cellcolor{ForestGreen!20}92.5\spl38.1~\y\y}
  & {\cellcolor{Maroon!20}26.8\spl\phantom{0}9.7~\n\n} \\

  & Coupling
  & {\cellcolor{ForestGreen!20}38.6\spl14.9~\y\y}
  & {\cellcolor{Maroon!20}230.0\spl\phantom{0}87.1~\n\n}
  & {\cellcolor{Maroon!20}71.7\spl27.0~\n\n}
  & {\cellcolor{Maroon!20}20.2\spl\phantom{0}7.8~\n\n} \\
\addlinespace

%% GROUP MODELS
Group
  & Means
  & {\cellcolor{ForestGreen!20}38.6\spl14.9~\y\y}
  & {\cellcolor{Maroon!20}311.4\spl109.5~\n\n}
  & {\cellcolor{Maroon!20}81.3\spl31.2~\n\n}
  & {\cellcolor{Goldenrod!20}30.2\spl13.0~\y\n} \\

  & Sampling
  & {\cellcolor{ForestGreen!20}38.6\spl14.9~\y\y}
  & {\cellcolor{Maroon!20}309.2\spl115.1~\n\n}
  & {\cellcolor{ForestGreen!20}91.2\spl38.4~\y\y}
  & {\cellcolor{ForestGreen!20}30.2\spl14.8~\y\y} \\

  & Coupling
  & {\cellcolor{ForestGreen!20}38.6\spl14.9~\y\y}
  & {\cellcolor{Maroon!20}285.7\spl104.0~\n\n}
  & {\cellcolor{ForestGreen!20}86.6\spl37.1~\y\y}
  & {\cellcolor{ForestGreen!20}29.4\spl14.8~\y\y} \\
\addlinespace

%% NODAL MODELS
Region
  & Means
  & {\cellcolor{ForestGreen!20}38.6\spl14.9~\y\y}
  & {\cellcolor{Goldenrod!20}303.7\spl103.0~\y\n}
  & {\cellcolor{Goldenrod!20}94.0\spl34.4~\y\n}
  & {\cellcolor{ForestGreen!20}31.0\spl13.6~\y\y} \\

  & Sampling
  & {\cellcolor{ForestGreen!20}38.6\spl14.9~\y\y}
  & {\cellcolor{Goldenrod!20}301.0\spl110.2~\y\n}
  & {\cellcolor{ForestGreen!20}90.4\spl37.4~\y\y}
  & {\cellcolor{ForestGreen!20}31.0\spl15.7~\y\y} \\

  & Coupling
  & {\cellcolor{ForestGreen!20}38.6\spl14.9~\y\y}
  & {\cellcolor{Goldenrod!20}296.2\spl108.6~\y\n}
  & {\cellcolor{ForestGreen!20}90.0\spl38.1~\y\y}
  & {\cellcolor{ForestGreen!20}30.0\spl15.1~\y\y} \\

\bottomrule
\end{tabular}

\vspace{3pt}
\begin{minipage}{0.88\linewidth}
\scriptsize
\textit{Notes.} 
\y = $p > 0.05$ and \n = $p \le 0.05$ under Welch’s $t$-test (means) and Levene’s test (variances).
Red/yellow/green shading indicates whether zero, one, or two statistical moments (mean and variance) match empirical data.
\end{minipage}

\end{table}

\begin{table}[!t]
\centering
\small
\setlength{\tabcolsep}{6pt}
%\caption{Spatial and temporal fidelity of synthetic nodal distributions relative to empirical data. Spatial fidelity is measured using the Jensen--Shannon divergence (JSD; lower is better). Temporal fidelity evaluates whether the model preserves the empirical dependence of shots and victims on dwell time. For each metric, Spearman's rank correlation $\rho$ is computed between time spent in a node and the corresponding outcome (shots or victims), separately for empirical and synthetic data. Temporal deviation is defined as $\Delta\rho = \rho_{\text{model}} - \rho_{\text{emp}}$, where values closer to zero indicate better preservation of empirical temporal dependencies. Best values per column are shown in bold.}
\caption{Fidelity of generated data compared to observed no-robot participant data. The numbers of generated and observed samples were 600 and 60, respectively. See notes for determination of fidelity.}
\label{tab:episode_spatial_temporal}
\begin{tabular}{ll
    S[table-format=1.3, table-column-width=1.4cm ]
    S[table-format=1.3, table-column-width=1.4cm]
    S[table-format=1.3, table-column-width=1.4cm]
    @{\hspace{6pt}}
    S[table-format=1.3, table-column-width=1.4cm]
    S[table-format=1.3, table-column-width=1.4cm]
    @{\hspace{6pt}}
    S[table-format=1.3, table-column-width=1.4cm]
    S[table-format=1.3, table-column-width=1.4cm]}
\toprule
    \multicolumn{2}{c}{} 
      & \multicolumn{3}{c}{\textbf{Spatial Fidelity, JSD}} 
      & \multicolumn{4}{c}{\textbf{Temporal Fidelity, Spearman $\rho$}} \\
\cmidrule(lr){3-5} \cmidrule(lr){6-9}
    \textbf{Pooling} 
      & \textbf{Variant}
      & \hdr{Time}
      & \hdr{Shots}
      & \hdr{Victims}
      & \hdr{$\rho(t,s)$}
      & \hdr{$\Delta\rho(t,s)$}
      & \hdr{$\rho(t,v)$}
      & \hdr{$\Delta\rho(t,v)$}\\
\midrule

\multicolumn{1}{c}{---}
  & \textit{Participants}
  & {---}
  & {---}
  & {---}
  & 0.515
  & {---}
  & 0.492
  & {---} \\
\addlinespace

Global
  & Means
  & 0.378
  & 0.638
  & 0.674
  & 0.704
  & +0.189
  & -0.088
  & -0.580 \\

  & Sampling
  & 0.379
  & 0.637
  & 0.689
  & 0.019
  & -0.496
  & -0.004
  & -0.496 \\

  & Coupling
  & 0.191
  & 0.492
  & 0.534
  & 0.949
  & +0.434
  & 0.834
  & +0.342 \\
\addlinespace

Group
  & Means
  & 0.158
  & 0.289
  & 0.223
  & 0.645
  & +0.130
  & 0.612
  & +0.120 \\

  & Sampling
  & 0.158
  & 0.184
  & 0.201
  & 0.345
  & -0.170
  & 0.459
  & -0.033 \\

  & Coupling
  & 0.037
  & 0.180
  & 0.204
  & 0.816
  & +0.301
  & 0.527
  & +0.035 \\
\addlinespace

Region
  & Means
  & \textbf{0.017}
  & 0.135
  & 0.208
  & 0.673
  & +0.158
  & 0.589
  & +0.097 \\

  & Sampling
  & 0.020
  & \textbf{0.079}
  & \textbf{0.180}
  & \textbf{0.490}
  & \textbf{-0.025}
  & \textbf{0.466}
  & \textbf{-0.026} \\

  & Coupling
  & 0.019
  & 0.102
  & 0.187
  & 0.664
  & +0.149
  & 0.530
  & +0.038 \\
\bottomrule
\end{tabular}

\vspace{3pt}
\begin{minipage}{0.95\linewidth}
\scriptsize
\textit{Notes.} 
JSD = Jensen–Shannon divergence; lower values indicate greater similarity between empirical and synthetic spatial distributions.  
For temporal fidelity, Spearman’s $\rho$ is computed between dwell time and event outcomes (shots, victims).  
Temporal deviation is defined as $\Delta\rho = \rho_{\text{model}} - \rho_{\text{emp}}$; values closer to zero indicate better preservation of empirical temporal structure.  
Boldface indicates the best-performing value in each column.
\end{minipage}
\end{table}

To determine which modeling choices best met these criteria, we evaluated multiple event-generation variants spanning different spatial and temporal resolutions. Spatial resolution was varied across three levels---region, group, and global---corresponding to increasingly pooled outcome statistics. Temporal generation was varied across three strategies: using participant means, sampling from moment-matched truncated distributions, and a coupled approach in which only dwell time was sampled and shots and victims were derived from participant mean rates. Crossing these dimensions yielded nine distinct event-generation variants.

Event-generation performance was evaluated using a leave-one-subset-out cross-validation protocol. The participant data were randomly ordered by episode and partitioned into five subsets. In each split, four subsets (80\%) were used to estimate outcome statistics, and the remaining subset (20\%) was used for testing. This procedure was repeated until each subset had served as the test set once. The results of this evaluation are shown in Tables~\ref{tab:episode_outcomes} and~\ref{tab:episode_spatial_temporal}. Table~\ref{tab:episode_outcomes} shows that, with respect to region occupancy, shots fired, and victims, only the region-level sampling and region-level coupling variants adequately matched participant means and variances. For dwell time, a variance mismatch relative to participant data was expected because episodes were capped at approximately 300~s, limiting long observations and reducing the participant variance. The simulator generated events were not subject to this cap and therefore exhibit greater variability, even when moment-matched at the region level. Table~\ref{tab:episode_spatial_temporal} further shows that the highest spatial and temporal fidelity was achieved using region-level pooling, typically with the sampling variant. The coupling variant performed comparably and did not reduce consistency in shot and victim rates, as indicated by positive $\Delta\rho$ values.

% ---------------------------------------------
\subsection{Robot Effects}
To evaluate robot effects, we compared simulator generated event outcomes with and without robot-effect modulation against the observed event sequences from human subjects in the robot-present condition. Evaluation focused on matching marginal outcome statistics, with differences in means and variances assessed using Welch’s unequal-variance $t$-tests and Levene’s tests, respectively. The results are shown in Table~\ref{tab:robot_effect}. Without robot-effect modulation, the region-level coupling model exhibits significant deviations in dwell time and shot outcomes, as indicated by \n~markers for both mean and variance. Introducing robot-effect modulation substantially improves agreement for shots and victims, yielding matched means and variances for all non-time outcomes. A variance discrepancy for dwell time remains, as expected due to the fixed episode duration, but the model accurately reproduces the participant mean dwell time. Overall, incorporating robot effects meaningfully improves the simulator’s ability to capture robot-induced behavioral changes.

\begin{table}[!t]
\centering
\small
\setlength{\tabcolsep}{7pt}

\caption{Generated versus participant outcomes with robot present. The numbers of generated and participant samples were 600 and 60, respectively. See notes for moment matching indicators.}
\label{tab:robot_effect}
\begin{tabular}{ll
    S[table-format=2.1]
    S[table-format=3.1]
    S[table-format=3.1]
    S[table-format=2.1]}
\toprule
\multicolumn{2}{c}{} &
\multicolumn{4}{c}{\textbf{Episode Outcomes: M ± SD with Moment-Matching Indicators}} \\
\cmidrule(lr){3-6}

\textbf{Pooling} & \textbf{Variant}
  & \hdr{Nodes}
  & \hdr{Time$^\dagger$}
  & \hdr{Shots}
  & \hdr{Victims} \\
\midrule

%% EMPIRICAL
\multicolumn{1}{c}{---}
  & \textit{Participants}
  & {\cellcolor{Gray!20}29.2\spl14.7~\s\s}
  & {\cellcolor{Gray!20}299.6\spl\hphantom{00}0.3~\s\s}
  & {\cellcolor{Gray!20}92.0\spl48.0~\s\s}
  & {\cellcolor{Gray!20}21.8\spl13.7~\s\s} \\
\addlinespace

%% MODELS
\multicolumn{1}{c}{Region}
  & Coupling
  & {\cellcolor{ForestGreen!20}29.2\spl14.7~\y\y}
  & {\cellcolor{Maroon!20}230.7\spl\hphantom{0}96.8~\n\n}
  & {\cellcolor{Maroon!20}67.8\spl35.5~\n\n}
  & {\cellcolor{ForestGreen!20}22.0\spl14.6~\y\y} \\

\multicolumn{1}{c}{Region}
  & Coupling + mod.
  & {\cellcolor{ForestGreen!20}29.2\spl14.7~\y\y}
  & {\cellcolor{Goldenrod!20}307.3\spl109.9~\y\n}
  & {\cellcolor{ForestGreen!20}80.7\spl39.3~\y\y}
  & {\cellcolor{ForestGreen!20}20.9\spl14.3~\y\y} \\

\bottomrule
\end{tabular}

\vspace{3pt}
\begin{minipage}{0.92\linewidth}
\scriptsize
\textit{Notes.} 
\y = $p > 0.05$ and \n = $p \le 0.05$ under Welch’s $t$-test (means) and Levene’s test (variances).
Red/yellow/green shading indicates whether zero, one, or two statistical moments (mean and variance) match participant data.
\end{minipage}

\end{table}

% ============================================
\newpage
\section{Policy Demonstration}\label{sec:policy_demo}
Having evaluated the individual components of our discrete-event simulator, we next demonstrate its suitability for rapid policy iteration and learning. For this demonstration, the intervention was carried out by two mobile robots jointly controlled by a single policy with the objective of minimizing the number of victims. To this end, we first examine how the simulator can be used to rapidly assess simple, hand-designed policies before introducing learning-based control.

\noindent \textbf{Policy Iteration.}
Table~\ref{tab:heuristics} summarizes outcomes for a small set of robot strategies evaluated under different mobility constraints, all generated within minutes of wall-clock time using the simulator. Strategies in which robots moved toward the shooter consistently resulted in fewer victims than other configurations, reflecting trends observed in participant studies. Allowing robots to move between floors further improved performance relative to configurations in which robot motion was restricted to a single floor. Smaller but consistent gains were also achieved through strategic placement alone, such as positioning robots in locations associated with greater reductions in victim rate. Although simple, these strategies demonstrate how surrogate simulation can yield actionable insights without requiring additional human-subject experiments.

\begin{table}[!h]
\centering
\small
\setlength{\tabcolsep}{15pt}

\caption{Generated victim outcomes for representative robot strategies under varying mobility constraints. Each value summarizes 600 simulated samples. See notes for details.}
\label{tab:heuristics}
\begin{tabular}{lcccc}
\toprule
& \multicolumn{2}{c}{\textbf{Single-Floor}} & 
\multicolumn{2}{c}{\textbf{Multi-Floor}} \\
\cmidrule(lr){2-3}
\cmidrule(lr){4-5}
\textbf{Robot Strategy}& \hdr{Victims} & \hdr{$\Delta$}& \hdr{Victims} & \hdr{$\Delta$}\\
\midrule
Not present & $31.15 \spl  11.26$  & -- & $31.15 \spl  11.26$  & --\\
Stay in initial position &$25.99 \spl   \hphantom{0}9.79$  &  $-16.6\%$ & $25.99 \spl  \hphantom{0}9.79$  &  $-16.6\%$\\
Move to low-impact region &$28.14 \spl  10.38$  &  $ -\hphantom{0}9.7\%$ & $28.14 \spl  10.38$  &  $ -\hphantom{0}9.7\%$\\
Move to high-impact region &$25.05 \spl   \hphantom{0}8.88$  &  $-19.6\%$ & $25.05 \spl   \hphantom{0}8.88$  &  $-19.6\%$\\
Move to shooter region &$ 20.75 \spl   \hphantom{0}9.59$  &  $-33.4\%$  & $\mathbf{17.58 \spl   \hphantom{0}8.90}$  &  $\mathbf{-43.6\%}$\\

\bottomrule
\end{tabular}

\vspace{6pt}
\begin{minipage}{0.92\linewidth}
\scriptsize
\textit{Notes.} All generated outcomes used region-level coupling for event outcome generation. Single-floor strategies restrict robots to their initial floor, whereas multi-floor strategies allow traversal between floors via stairwells. $\Delta$ denotes the percent difference relative to the no-robot baseline. High- and low-impact regions are defined by the empirical sensitivity of victim rate to robot presence ($d\dot{V}/dR$).
\end{minipage}
\end{table}

\begin{comment}
\toprule
& \multicolumn{2}{c}{\textbf{Single-Floor}} & 
\multicolumn{2}{c}{\textbf{Multi-Floor}} \\
\cmidrule(lr){2-3}
\cmidrule(lr){4-5}
\textbf{Robot Strategy}& \hdr{Victims} & \hdr{$\Delta$}& \hdr{Victims} & \hdr{$\Delta$}\\
\midrule
\end{comment}

\noindent \textbf{Policy Learning.}
Next, we used reinforcement learning (RL) to learn a pursuit-based intervention strategy corresponding to the best-performing policy in Table~\ref{tab:heuristics}. The discrete-event simulator was embedded within a Double Deep Q-Network (DDQN)~\cite{van2016deep}, using a simple two-layer multilayer perceptron (MLP) to approximate the joint action-value function~\cite{bishop2006prml}. To maintain a fixed-dimensional action space, each robot used a discrete action set sized to the maximum number of valid movements from any region, with invalid actions masked. Actions were executed until completion or episode termination, while the shooter progressed independently using the DES. Observations consisted of an action-conditioned feature vector. For each available robot action, the vector encoded the resulting robot–shooter distance if that action were executed from the current state. This representation provides local, one-step structural information without incorporating any explicit prediction or rollout of shooter behavior. Each completed agent action received a reward defined as $\mathcal{R} = -\alpha(d_1 + d_2)$, where $\alpha$ is a normalizing constant and $d_1$ and $d_2$ denote the graph distances from the shooter to robot~1 and robot~2, respectively. Full implementation details, including network configuration, action masking, training procedures, and hyperparameter settings, are provided on GitHub.$^1$

Training to convergence required approximately 15{,}000 episodes, corresponding to less than nine hours of wall-clock time. During training, the policy was learned by iteratively interacting with the discrete-event simulator, receiving rewards, and updating policy parameters. By contrast, conducting an equivalent number of episodes using human participants in VR---each lasting five minutes---would require approximately 52.1 days of continuous experimentation. When evaluated under the same conditions as Table~\ref{tab:heuristics}, the learned policy produced $19.34 \spl 9.11$ victims, corresponding to a $37.9\%$ reduction relative to the no-robot baseline. Although this performance does not exceed the best-performing hand-designed policy, it demonstrates that stable and effective policy learning is achievable using the discrete-event simulator and suggests that further gains will require policies that reason beyond purely reactive behavior.

% ============================================
\section{Conclusion}

This paper develops a discrete-event simulator as a computationally efficient surrogate for time-consuming and expensive human-subject experiments, enabling behavioral analysis and policy development at time scales that would otherwise be infeasible. Comparison to baselines demonstrates the value of each model component. In particular, the shooter transition model, trained using semantically and topologically informed features, significantly outperforms baseline heuristics. The shooter event model, calibrated with participant data from virtual reality, generates event outcomes that match key statistical moments while preserving the spatial and temporal structure observed in participant behavior. 

There are several limitations to this work. First, the empirical distributions used for event sampling depend on contextual factors---such as time of day, building occupancy, and shooter armament---that were not varied in our study. Addressing these dependencies will require additional human-subject data collected under systematically manipulated conditions. Second, all empirical data were obtained in a single environment modeled after Columbine High School. Although the methodology is fully reproducible in other environments, and the transition model demonstrated strong generalization to real-shooter trajectories with unseen graph topologies, broader evaluation will require acquiring empirical data from additional layouts to support stronger claims of generalizability. Third, while the policy learned using the surrogate simulator was stable, the learned policy has not yet been validated in VR. Importantly, the primary contribution of this work lies not in the learned policy itself, but in the methods developed and used to enable such policies to be learned. 

Finally, this work involves modeling sensitive scenarios related to school violence. All human-subject data were collected under an Institutional Review Board (IRB)-approved protocol and classified as minimal risk under U.S. federal guidelines (45~C.F.R.~\S~46.102)~\cite{ecfr45cfr46}. The data collection was non-deceptive, participants were informed of the nature of the task, and safeguards were in place to allow withdrawal and monitor potential distress. All data were de-identified prior to analysis and release. While this work is intended to support the evaluation of defensive interventions, we recognize the potential for misuse and therefore limit the release of data and code.

Overall, this work establishes a principled high-to-mid fidelity workflow in which human-subject data informs a discrete-event surrogate that enables scalable policy learning. The next steps will be to assess learned policies with human subjects in virtual reality, completing the loop between simulation-based learning and human-subject evaluation. More broadly, this framework provides a foundation for future research on autonomous mitigation in safety-critical environments. 

% ============================================

\section*{Acknowledgments}
This material is based upon work supported by the National Science Foundation under Grant No.~IIS-2045146. Any opinions, findings, and conclusions or recommendations expressed in this material are those of the author(s) and do not necessarily reflect the views of the National Science Foundation. Portions of this paper were edited and refined with assistance from generative AI. The authors reviewed and take responsibility for all content.

% ============================================
\bibliographystyle{scsproc}
\bibliography{ref}
\end{document}